\documentclass{article}
\usepackage{graphicx} 
\usepackage{amsmath,amssymb,amsfonts}
\usepackage{booktabs}
\usepackage{array}
\usepackage{listings}
\usepackage{xcolor}
\usepackage[margin=1in]{geometry}
\usepackage{float}
\usepackage{algorithm}
\usepackage{algpseudocode}
\usepackage{caption}
\usepackage{subcaption}
\usepackage[hidelinks]{hyperref} 
\usepackage{xurl}
\usepackage{parskip}
\title{GPU-Accelerated INT8 Quantization for KV Cache Compression in Large Language Models}
\author{Maanas Taneja, Purab Shingvi}
\date{December 2025}

\begin{document}

\maketitle

\begin{abstract}
The key-value (KV) cache in large language models presents a significant memory bottleneck during inference, growing linearly with sequence length and often exceeding the memory footprint of model weights themselves. We implement and evaluate GPU-accelerated INT8 quantization for KV cache compression, achieving 4$\times$ memory reduction with minimal accuracy degradation. We develop four CUDA kernel variants---naive, tiled, coarsened, and vectorized---and benchmark them across realistic workload sizes up to 1 billion elements. Our vectorized kernel achieves up to 1,694$\times$ speedup over CPU baselines while maintaining reconstruction error below 0.004 and attention score error below 0.1 even for 8K-dimensional heads. These results demonstrate that INT8 quantization provides a practical approach for reducing memory pressure in LLM inference with negligible computational overhead (6--58ms) and minimal impact on downstream model behavior. Code available at: \url{https://github.com/MaanasTaneja/cuda-kv-cache-compression}
\end{abstract}

\section{Introduction}
Modern large language models (LLMs) based on the Transformer architecture have achieved impressive results across a wide range of tasks, from text generation to code completion. However, deploying these models efficiently presents a significant challenge: memory consumption during inference.

A key contributor to this memory bottleneck is the key-value (KV) cache. During autoregressive text generation, the model produces one token at a time, and at each step it must attend to all previously generated tokens. To avoid redundantly recomputing the key and value projections for past tokens, these values are stored in a cache. While this speeds up inference, the cache grows linearly with sequence length---for long contexts (32k, 128k tokens or more), the KV cache can consume tens of gigabytes of memory, sometimes exceeding the memory required by the model weights themselves.

This memory pressure has practical consequences: it limits the maximum context length, forces smaller batch sizes (reducing throughput), and increases serving costs on expensive GPU hardware. Quantization offers a promising solution to this problem. The idea is simple: instead of storing cached keys and values in 32-bit floating point (FP32), we compress them to 8-bit integers (INT8). This yields a 4$\times$ reduction in memory footprint. When the values are needed for attention computation, we dequantize them back to FP32. The tradeoff is a small loss in numerical precision due to rounding.

In this project, we implement and evaluate INT8 quantization and dequantization for KV cache compression. We use per-channel quantization, where each dimension of the key vectors gets its own scale factor, allowing us to preserve precision across dimensions with different value ranges. We implement a CPU baseline and four GPU kernel variants---naive, tiled, coarsened, and vectorized---to explore performance optimization strategies on parallel hardware.

To evaluate our implementation, we measure:
\begin{itemize}
    \item \textbf{Performance}: How fast are the quantization and dequantization operations? What speedup do the GPU kernels achieve over the CPU?
    \item \textbf{Reconstruction error}: How closely does the dequantized matrix match the original? We report L2 error and maximum absolute error.
    \item \textbf{Attention score error}: Does quantization affect downstream computation? We measure the mean absolute difference between attention dot products computed with original and reconstructed keys.
\end{itemize}

\section{Related Work}

\subsection{KV Cache Optimization}
Recent work has explored various approaches to reducing KV cache memory consumption. \textbf{FlashAttention}~\cite{dao2022flashattention} and \textbf{FlashAttention-2}~\cite{dao2023flashattention2} optimize attention computation through kernel fusion and memory hierarchy exploitation but do not directly compress the cache. \textbf{PagedAttention}~\cite{kwon2023efficient}, implemented in the vLLM serving system, applies paging techniques to reduce memory fragmentation but maintains full precision storage.

\subsection{Quantization Techniques}
Quantization has emerged as a promising direction for LLM compression. \textbf{LLM.int8()}~\cite{dettmers2022llm} demonstrates that 8-bit quantization can be applied to model weights and activations with minimal accuracy loss. \textbf{SmoothQuant}~\cite{xiao2022smoothquant} addresses activation quantization challenges through per-channel scaling. More recent work explores extreme quantization: \textbf{KIVI}~\cite{liu2024kivi} proposes 2-bit asymmetric quantization specifically for KV caches, while \textbf{KVQuant}~\cite{hooper2024kvquant} achieves sub-4-bit precision through learned quantization parameters.

\subsection{LLM Inference Systems}
Production inference frameworks have begun incorporating memory optimization techniques. \textbf{FlexGen}~\cite{sheng2023flexgen} explores offloading strategies for memory-constrained scenarios. \textbf{TensorRT-LLM} and similar systems~\cite{pope2022efficiently} integrate quantization into optimized inference pipelines. However, these systems often treat quantization as a black box, providing limited insight into the performance characteristics of different kernel implementations.

\subsection{Our Contribution}
Our work focuses on the implementation and systematic evaluation of per-channel INT8 quantization for KV cache compression. Unlike prior work that primarily evaluates end-to-end model accuracy, we provide detailed analysis of kernel-level performance characteristics across multiple GPU optimization strategies. We demonstrate that vectorized memory operations provide consistent speedups for this memory-bound workload, while traditional optimizations like shared memory tiling offer limited benefit. Our analysis provides practical guidance for implementing efficient quantization kernels in production LLM serving systems.

\section{Background}

\subsection{Transformer Attention \& KV Caching}

The Transformer architecture relies on a mechanism called {self-attention, which allows each token in a sequence to attend to all other tokens. For a given input, the model computes three projections: queries ($Q$), keys ($K$), and values ($V$). The attention output is computed as:

\begin{equation}
\text{Attention}(Q, K, V) = \text{softmax}\left(\frac{QK^T}{\sqrt{d_k}}\right)V
\label{eq:attention}
\end{equation}

where $d_k$ is the dimension of the key vectors.

While producing one token at a time, the model must compute attention between the current token's query and the keys of all previous tokens. Naively, this would require recomputing the key and value projections for every past token at each generation step, a very expensive operation.

The KV cache solves this by storing the key and value projections from previous tokens. At each new generation step, only the current token's $K$ and $V$ are computed and appended to the cache. This transforms the complexity from quadratic to linear in sequence length per step, but introduces a memory cost: the cache must store all past keys and values.

\subsection{Memory Bottleneck in LLM Inference}

For a model with $L$ layers, $H$ attention heads, head dimension $d$, and sequence length $T$, the KV cache stores:

\begin{equation}
\text{KV cache size} = 2 \times L \times H \times d \times T \times \text{(bytes per element)}
\label{eq:kv_size}
\end{equation}

\begin{table}[H]
\centering
\caption{Example KV cache memory calculation}
\label{tab:memory_example}
\begin{tabular}{@{}ll@{}}
\toprule
\textbf{Parameter} & \textbf{Value} \\
\midrule
Layers ($L$) & 32 \\
Heads ($H$) & 32 \\
Head dimension ($d$) & 128 \\
Sequence length ($T$) & 131,072 \\
Precision & FP32 (4 bytes) \\
\midrule
\textbf{Total memory} & $\approx$137 GB \\
\bottomrule
\end{tabular}
\end{table}

Even with FP16, this is nearly 70 GB, exceeding the memory of most GPUs. This memory pressure limits batch sizes, restricts context lengths, and increases deployment costs.

\subsection{Quantization Fundamentals}

Quantization reduces memory by representing values in fewer bits. The core idea is to map floating point values to a smaller set of discrete integer levels.

For linear quantization to INT8, a floating point value $x$ is mapped to an 8-bit integer $x_q$ using:

\begin{equation}
x_q = \text{round}\left(\frac{x}{s}\right)
\label{eq:quantize}
\end{equation}

where $s$ is a scale factor that maps the floating point range to the integer range $[-127, 127]$.

To recover the approximate original value (\textbf{dequantization}):

\begin{equation}
\hat{x} = x_q \times s
\label{eq:dequantize}
\end{equation}

The reconstruction $\hat{x}$ differs from the original $x$ due to rounding error. This error is bounded by $\frac{s}{2}$ - half the quantization step size.

\textbf{Per-channel quantization} uses a separate scale for each dimension (column) of the matrix:

\begin{equation}
s_d = \frac{\max_t |K[t, d]|}{127}
\label{eq:scale}
\end{equation}

This allows dimensions with different value ranges to each utilize the full INT8 range, improving precision compared to a single global scale.

\section{Method}

\subsection{Per-Channel INT8 Quantization}

We apply per-channel quantization along the head dimension of the KV cache. For a key matrix $K$ with shape $(T, D)$ where $T$ is the number of tokens and $D$ is the head dimension, we compute one scale factor per column.

\subsection{Scale Computation}

For each dimension $d \in [0, D)$, the scale is computed as:

\begin{equation}
s_d = \frac{\max_{t \in [0, T)} |K[t, d]|}{127}
\label{eq:scale_compute}
\end{equation}

This requires a single pass over each column to find the maximum absolute value. The scales are stored as FP32 values, adding negligible memory overhead ($D$ floats compared to $T \times D$ elements in the cache).

The scale computation algorithm is presented in Algorithm~\ref{alg:compute_scales}.

\begin{algorithm}[H]
\caption{Compute Scales}
\label{alg:compute_scales}
\begin{algorithmic}[1]
\Require $K$: FP32 matrix of shape $T \times D$
\Ensure $\text{scales}$: FP32 array of length $D$
\For{$d = 0$ to $D-1$}
    \State $\text{max\_abs} \gets 0$
    \For{$t = 0$ to $T-1$}
        \If{$|K[t, d]| > \text{max\_abs}$}
            \State $\text{max\_abs} \gets |K[t, d]|$
        \EndIf
    \EndFor
    \State $\text{scales}[d] \gets \text{max\_abs} / 127$
\EndFor
\end{algorithmic}
\end{algorithm}

\subsection{Quantization and Dequantization Operations}

\textbf{Quantization} converts each FP32 element to INT8:

\begin{equation}
K_{\text{int8}}[t, d] = \text{round}\left(\frac{K[t, d]}{s_d}\right)
\label{eq:quant_op}
\end{equation}

The result is clamped to $[-127, 127]$ to handle any numerical edge cases.

\textbf{Dequantization} recovers approximate FP32 values:

\begin{equation}
\hat{K}[t, d] = K_{\text{int8}}[t, d] \times s_d
\label{eq:dequant_op}
\end{equation}

The round-trip introduces quantization error. For a value $x$ with scale $s$:

\begin{equation}
\text{error} = |x - \hat{x}| \leq \frac{s}{2}
\label{eq:error_bound}
\end{equation}

This error propagates to attention computation, but as we measure empirically, the impact on attention scores is typically small.

\section{Implementation}
\subsection{Data Structures}

We define two matrix structures to hold original and quantized data:

\begin{lstlisting}[language=C, caption={Matrix data structures}]
typedef struct {
    int rows;      // T (number of tokens)
    int columns;   // D (head dimension)
    float* data;   // row-major storage
} FP32Matrix;

typedef struct {
    int rows;
    int columns;
    int8_t* data;  // row-major storage
} INT8Matrix;
\end{lstlisting}

The quantized matrix uses 4$\times$ less memory than the original.

\subsection{CPU Baseline}

The CPU implementation provides a correctness reference and performance baseline.

\subsubsection{Scale Computation}

\begin{lstlisting}[language=C, caption={CPU scale computation}]
void compute_scales(FP32Matrix* K, float* scales) {
    for (int d = 0; d < K->columns; d++) {
        float max_abs = 0.0f;
        for (int t = 0; t < K->rows; t++) {
            float val = fabsf(K->data[t * K->columns + d]);
            if (val > max_abs) max_abs = val;
        }
        scales[d] = max_abs / 127.0f;
    }
}
\end{lstlisting}

\subsubsection{Quantization}

\begin{lstlisting}[language=C, caption={CPU quantization}]
void quantize_matrix(FP32Matrix* K, INT8Matrix* K_int8, float* scales) {
    for (int t = 0; t < K->rows; t++) {
        for (int d = 0; d < K->columns; d++) {
            float val = K->data[t * K->columns + d];
            int q = (int)roundf(val / scales[d]);
            // Clamp to INT8 range
            if (q > 127) q = 127;
            if (q < -127) q = -127;
            K_int8->data[t * K->columns + d] = (int8_t)q;
        }
    }
}
\end{lstlisting}

\subsubsection{Dequantization}

\begin{lstlisting}[language=C, caption={CPU dequantization}]
void dequantize_matrix(INT8Matrix* K_int8, FP32Matrix* K_recon, 
                       float* scales) {
    for (int t = 0; t < K_int8->rows; t++) {
        for (int d = 0; d < K_int8->columns; d++) {
            int8_t q = K_int8->data[t * K_int8->columns + d];
            K_recon->data[t * K_int8->columns + d] = (float)q * scales[d];
        }
    }
}
\end{lstlisting}

\subsection{GPU Kernel Variants}

We implement four CUDA kernel variants with increasing levels of optimization. All kernels perform the same computation but differ in how they map work to threads and utilize GPU memory hierarchy.

\subsubsection{Naive Kernel}

The naive approach assigns one thread per matrix element. Each thread independently loads its scale, performs the computation, and writes the result.

\begin{lstlisting}[language=C, caption={Naive CUDA kernel for quantization}]
__global__ void quantize_naive(float* K, int8_t* K_int8, 
                               float* scales, int T, int D) {
    int t = blockIdx.y * blockDim.y + threadIdx.y;  // row
    int d = blockIdx.x * blockDim.x + threadIdx.x;  // column
    
    if (t < T && d < D) {
        float val = K[t * D + d];
        float s = scales[d];
        int q = __float2int_rn(val / s);
        q = max(-127, min(127, q));
        K_int8[t * D + d] = (int8_t)q;
    }
}
\end{lstlisting}

\textbf{Characteristics:}
\begin{itemize}
    \item Simple \& easy to understand
    \item Memory coalescing access.
    \item Redundant scale loads -multiple threads loading the same scale
\end{itemize}

\subsubsection{Tiled Kernel}

The tiled kernel uses shared memory to cache scales, reducing redundant global memory accesses.

\begin{lstlisting}[language=C, caption={Tiled CUDA kernel for quantization}]
__global__ void quantize_tiled(float* K, int8_t* K_int8,
                               float* scales, int T, int D) {
    __shared__ float s_scales[TILE_DIM];
    
    int t = blockIdx.y * blockDim.y + threadIdx.y;
    int d = blockIdx.x * blockDim.x + threadIdx.x;
    
    if (threadIdx.y == 0 && d < D) {
        s_scales[threadIdx.x] = scales[d];
    }
    __syncthreads();
    
    if (t < T && d < D) {
        float val = K[t * D + d];
        float s = s_scales[threadIdx.x];
        int q = __float2int_rn(val / s);
        q = max(-127, min(127, q));
        K_int8[t * D + d] = (int8_t)q;
    }
}
\end{lstlisting}

\textbf{Characteristics:}
\begin{itemize}
    \item Scales loaded once per block, then reused from fast shared memory
    \item Requires thread synchronization (\texttt{\_\_syncthreads()})
    \item Better memory efficiency for scale access
\end{itemize}

\subsubsection{Coarsened Kernel}

Thread coarsening assigns multiple elements to each thread, amortizing thread management overhead and improving instruction-level parallelism.

\begin{lstlisting}[language=C, caption={Coarsened CUDA kernel for quantization}]
__global__ void quantize_coarsened(float* K, int8_t* K_int8,
                                   float* scales, int T, int D) {
    int d = blockIdx.x * blockDim.x + threadIdx.x;
    
    if (d < D) {
        float s = scales[d];

        for (int t = blockIdx.y; t < T; t += gridDim.y) {
            float val = K[t * D + d];
            int q = __float2int_rn(val / s);
            q = max(-127, min(127, q));
            K_int8[t * D + d] = (int8_t)q;
        }
    }
}
\end{lstlisting}

\textbf{Characteristics:}
\begin{itemize}
    \item Each thread loads its scale once, processes many elements
    \item Reduces total number of threads needed
    \item Better register utilization
\end{itemize}

\subsubsection{Vectorized Kernel}

The vectorized kernel uses vector load/store instructions (\texttt{float4}, \texttt{char4}) to process multiple elements per memory transaction, improving memory bandwidth utilization.

\begin{lstlisting}[language=C, caption={Vectorized CUDA kernel for quantization}]
__global__ void quantize_vectorized(float* K, int8_t* K_int8,
                                    float* scales, int T, int D) {
    int t = blockIdx.y * blockDim.y + threadIdx.y;
    int d4 = (blockIdx.x * blockDim.x + threadIdx.x) * 4;
    
    if (t < T && d4 + 3 < D) {
        // Vector load: 4 floats at once
        float4 vals = *reinterpret_cast<float4*>(&K[t * D + d4]);
        float4 s = *reinterpret_cast<float4*>(&scales[d4]);
        
        char4 q;
        q.x = (int8_t)max(-127, min(127, __float2int_rn(vals.x / s.x)));
        q.y = (int8_t)max(-127, min(127, __float2int_rn(vals.y / s.y)));
        q.z = (int8_t)max(-127, min(127, __float2int_rn(vals.z / s.z)));
        q.w = (int8_t)max(-127, min(127, __float2int_rn(vals.w / s.w)));
        
        *reinterpret_cast<char4*>(&K_int8[t * D + d4]) = q;
    }
}
\end{lstlisting}

\textbf{Characteristics:}
\begin{itemize}
    \item 4$\times$ fewer memory transactions
    \item Better memory coalescing
    \item Requires $D$ to be divisible by 4 (or handle edge cases)
    \item Highest bandwidth utilization
\end{itemize}

\section{Tests}

\subsection{Hardware Configurations}

\begin{table}[H]
\centering
\caption{Hardware configuration}
\label{tab:hardware}
\begin{tabular}{@{}ll@{}}
\toprule
\textbf{Component} & \textbf{Specification} \\
\midrule
Server & Dell PowerEdge R740 \\
GPU & NVIDIA Tesla T4 (16 GB GDDR6) \\
CPU & 2× Intel Xeon Gold 6148 @ 2.40 GHz (40 cores) \\
RAM & 96 GB \\
OS & Ubuntu \\
CUDA Version & 12.0 \\
\bottomrule
\end{tabular}
\end{table}

\subsection{Test Configurations}
We benchmark across eight test cases representing different workload sizes, as shown in Table~\ref{tab:test_cases}.

\begin{table}[H]
\centering
\caption{Test configurations for benchmarking}
\label{tab:test_cases}
\begin{tabular}{@{}llll@{}}
\toprule
\textbf{Test Case} & \textbf{Tokens ($T$)} & \textbf{Head Dim ($D$)} & \textbf{Description} \\
\midrule
Small & 2,048 & 128 & Minimal test \\
Medium & 16,384 & 256 & Development testing \\
Large & 65,536 & 256 & Extended context \\
Very Large & 131,072 & 256 & Long context \\
Realistic Small & 131,072 & 1,024 & Small LLM workload \\
Realistic Medium & 131,072 & 2,048 & Medium LLM workload \\
Realistic Large & 131,072 & 4,096 & Large LLM workload \\
Realistic V. Large & 131,072 & 8,192 & Very large LLM workload \\
\bottomrule
\end{tabular}
\end{table}

\section{Evaluation}

\subsection{Performance Results}

Figure 1 shows GPU speedup across all test configurations. The vectorized kernel consistently outperforms other variants, achieving 1,694× speedup on the largest workload (131K × 8K). Speedup improves with problem size as larger workloads better amortize GPU launch overhead.

\begin{figure}[H]
\centering
\includegraphics[width=0.6\textwidth]{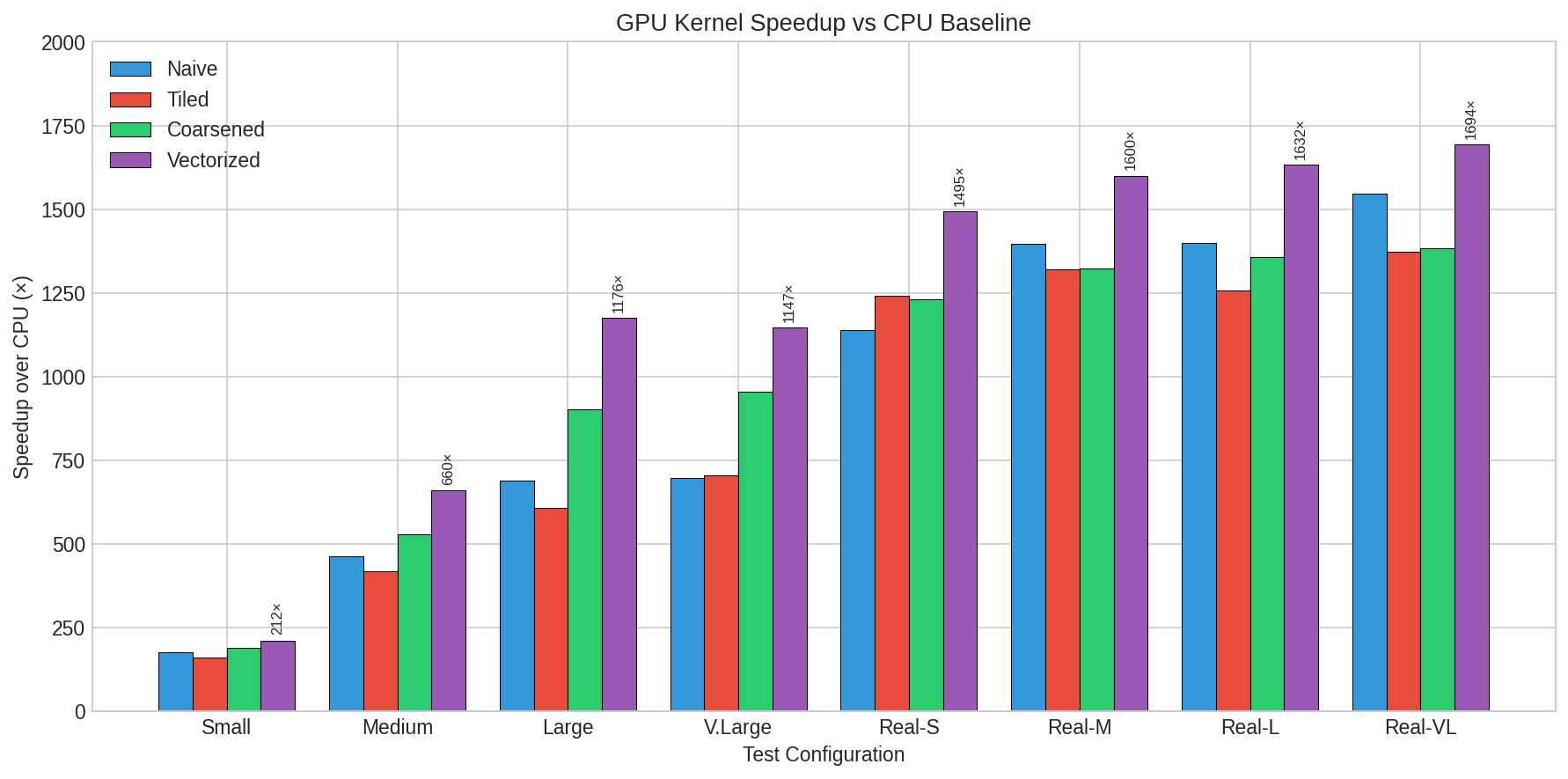}
\caption{GPU kernel speedup comparison across all test configurations.}
\label{fig:speedup}
\end{figure}

Interestingly, the tiled kernel does not outperform the naive kernel despite using shared memory. This suggests the scales array is small enough to remain in L2 cache, making explicit caching unnecessary.

Figure 2 shows execution time on a log-log scale. The gap between CPU and GPU spans three orders of magnitude. Even for the largest configuration (1 billion elements), GPU kernels complete in under 50ms compared to 79 seconds on CPU.

\begin{figure}[H]
\centering
\includegraphics[width=0.6\textwidth]{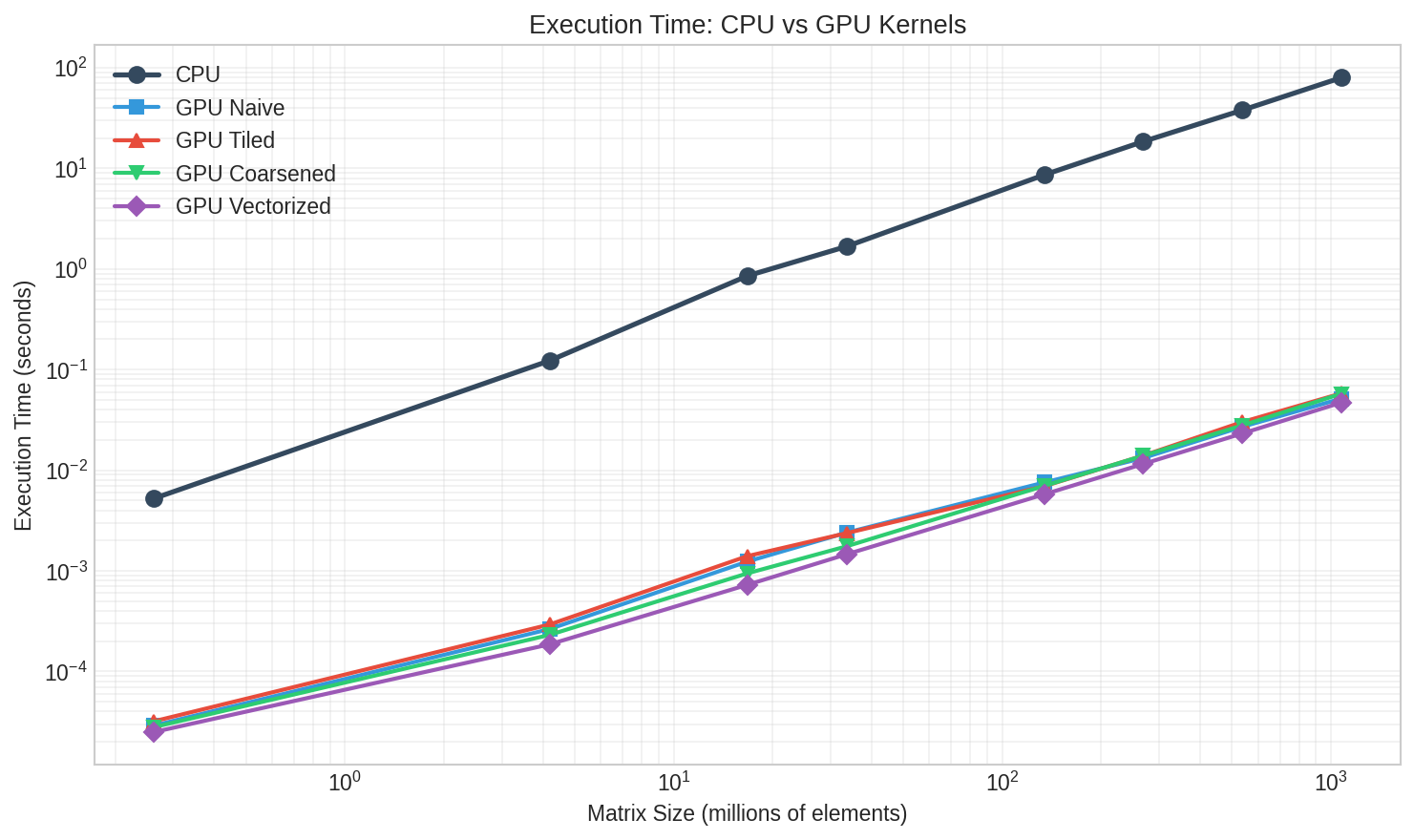}
\caption{Execution Time: CPU vs GPU}
\label{fig:errors}
\end{figure}

For realistic LLM workloads (Figure 3), all GPU kernels complete in 6–58 milliseconds, demonstrating that quantization overhead is negligible compared to actual attention computation.

\begin{figure}[H]
\centering
\includegraphics[width=0.6\textwidth]{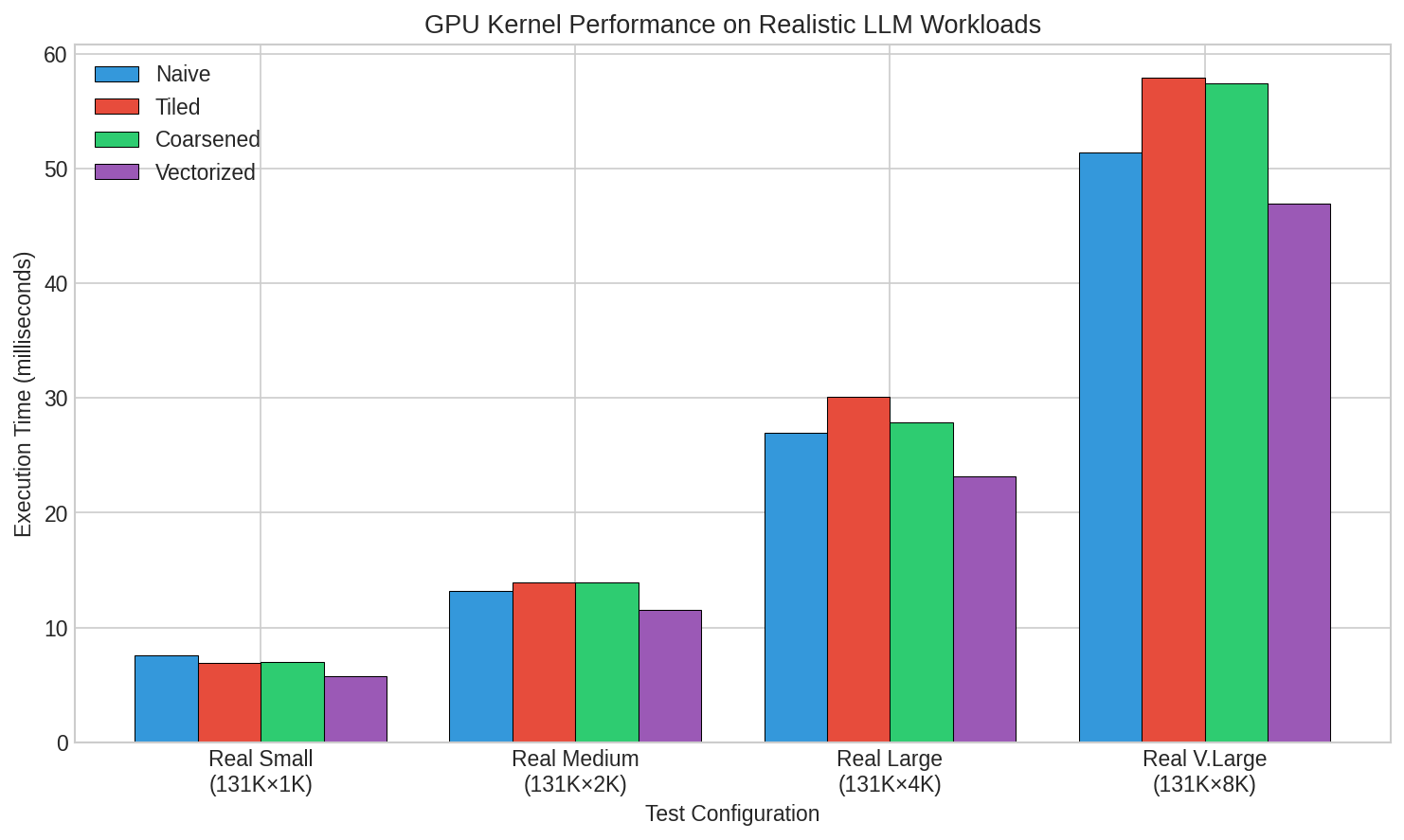}
\caption{GPU Performance on realistic LLM loads.}
\label{fig:errors}
\end{figure}

\subsection{Reconstruction Error}

All GPU kernels produce identical results to the CPU implementation, confirming correctness.

The maximum absolute error is constant at 0.00394 across all configurations (Figure 4, left). This matches the theoretical bound: with input values in [-1, 1] and 255 quantization levels, max error = 1/(2×127) equaling roughly 0.00394.

\begin{figure}[H]
\centering
\includegraphics[width=0.6\textwidth]{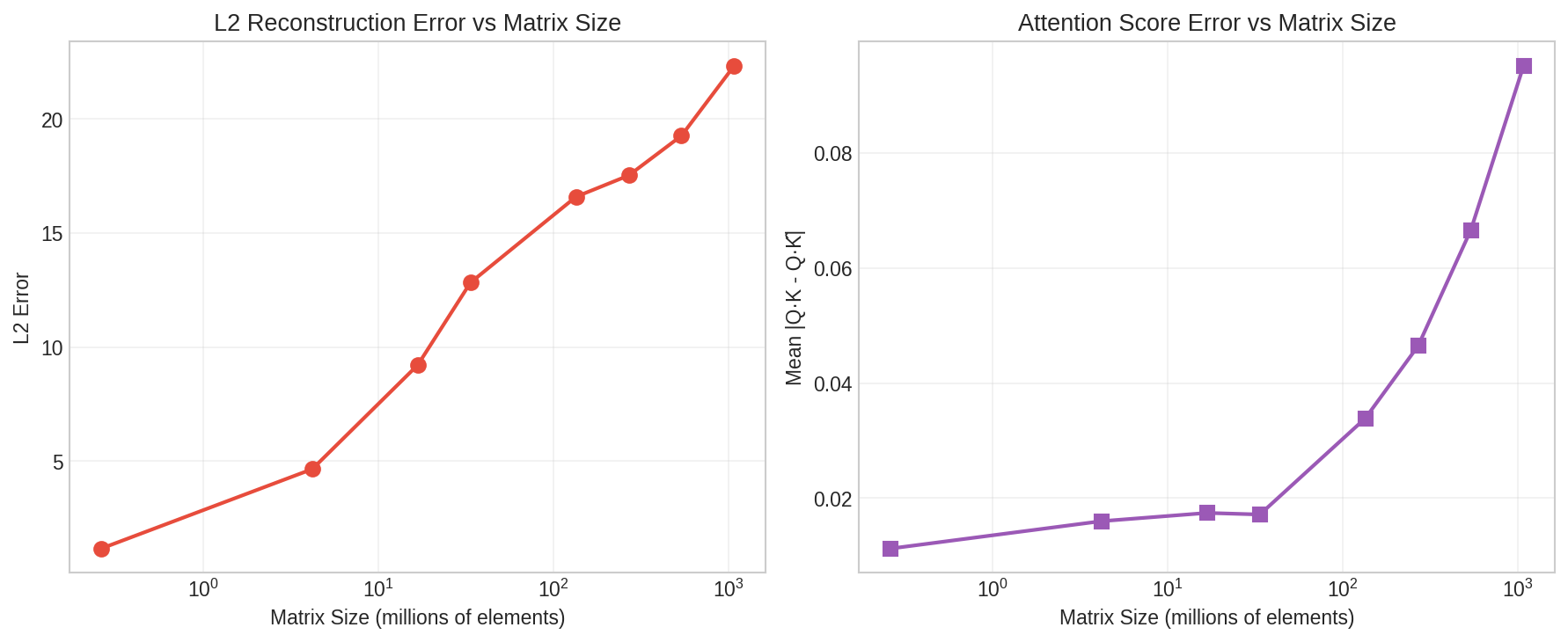}
\caption{L2 Reconstruction \& Attention Score Error vs Matrix Size.}
\label{fig:errors}
\end{figure}

L2 error grows with matrix size because it sums errors across all elements, this is expected and does not indicate degraded per-element precision.

\subsection{Attention Score Error}
The attention score error measures whether quantization affects downstream computation. Figure 4 (right) show that error scales approximately as $\sqrt{D}$ with head dimension.

Crucially, even for the largest configuration (D=8192), mean attention error is only 0.095. Given that attention scores typically span several orders of magnitude before softmax, this error is unlikely to meaningfully alter attention distributions.

\subsection{Discussion: Kernel Behavior and Architectural Implications}

A key observation from the performance results is that the naive kernel performs competitively with, and in some cases slightly better than, the shared-memory tiled kernel. This outcome appears counterintuitive at first, since tiling is often expected to improve performance by reducing global memory traffic. However, this behavior is explained by the memory-access characteristics of the quantization workload.

KV cache quantization is a strictly element-wise operation with no data reuse across threads. Each element is read once, scaled, quantized, and written back. As a result, the computation is overwhelmingly memory-bound, with extremely low arithmetic intensity. In such cases, performance is dominated by global memory bandwidth rather than arithmetic throughput. Because the naive kernel already issues fully coalesced global memory loads and stores along the column dimension, it effectively saturates memory bandwidth without additional optimization.

The tiled kernel introduces shared memory but does not benefit from reuse, since each element is consumed by exactly one thread. Moreover, explicit shared-memory staging introduces additional instructions and synchronization overhead. Although this overhead is small, it offsets any potential benefit, resulting in performance that closely matches, but does not surpass, the naive implementation. The results therefore reinforce the principle that shared memory is beneficial primarily when data reuse exists, not simply as a default optimization.

Thread coarsening improves performance modestly by reducing instruction overhead and improving instruction-level parallelism within each thread. By allowing each thread to process multiple contiguous columns, coarsening reduces loop and address-computation overhead while maintaining coalesced memory access. However, because memory bandwidth remains the dominant bottleneck, the gains from coarsening alone are limited and plateau quickly as the workload scales.

In contrast, the vectorized kernel delivers consistent and substantial performance improvements across all problem sizes. Vectorization reduces the number of memory transactions by loading and storing multiple elements per instruction, effectively increasing memory throughput without changing the algorithmic structure. This optimization directly targets the primary bottleneck of the workload. Unlike shared memory or coarsening, vectorization improves effective bandwidth utilization rather than reducing arithmetic overhead, making it uniquely well-suited for this access pattern.

Taken together, these results confirm that KV cache quantization is fundamentally a memory-bound operation. Once memory accesses are fully coalesced, further optimization requires reducing the number of memory instructions rather than restructuring computation. Vectorization achieves this goal, while tiling and aggressive coarsening offer diminishing returns. Beyond vectorization, additional kernel-level optimizations are unlikely to yield meaningful gains without changes to the data layout or precision format.

\begin{figure}[H]
\centering
\includegraphics[width=0.6\textwidth]{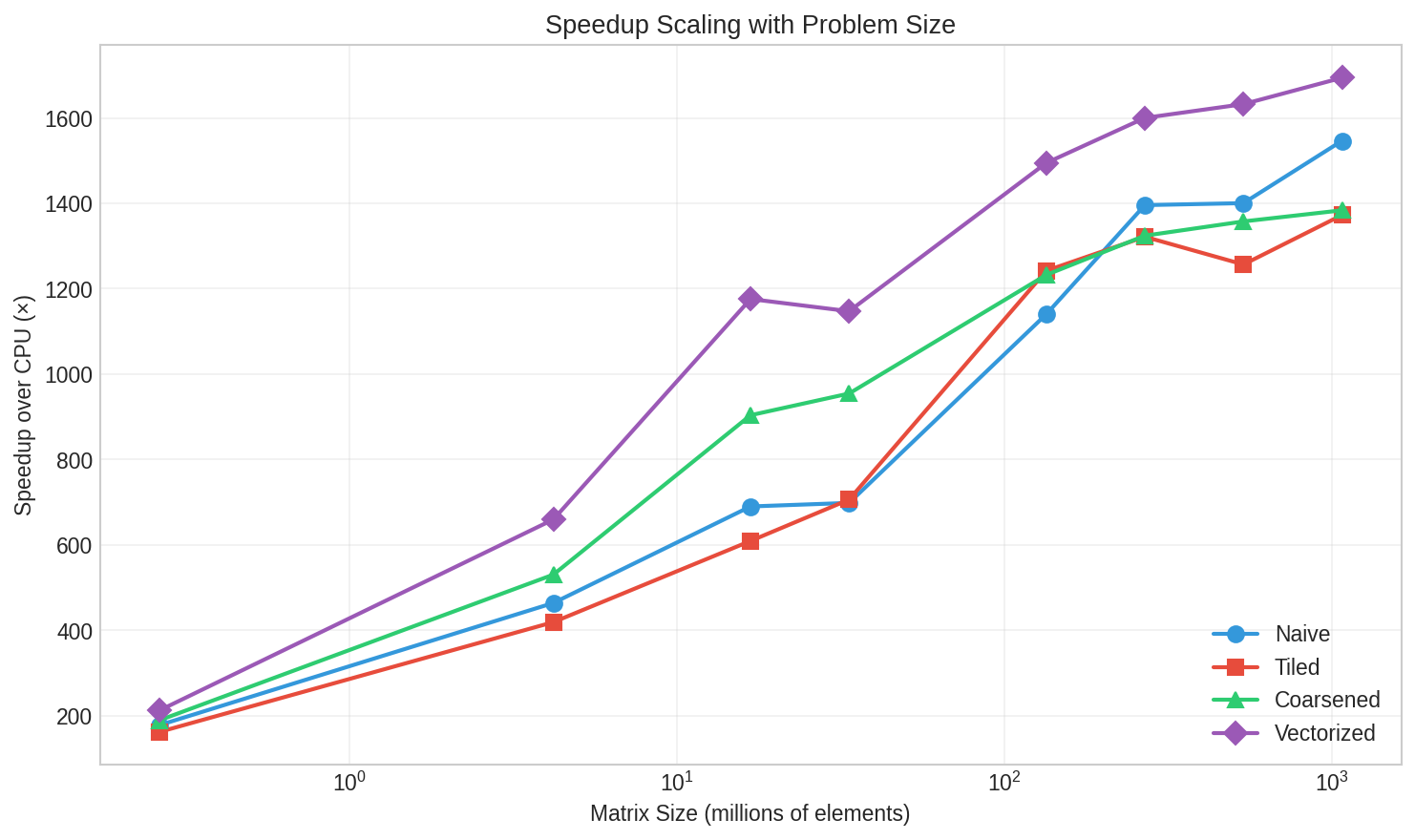}
\caption{Speed Scale Ups vs Problem Size.}
\label{fig:errors}
\end{figure}

\subsection{Test Case Design and Rationale}

To ensure that our evaluation reflects both theoretical scalability and real-world deployment scenarios, we carefully selected a diverse set of test configurations that span from synthetic micro-benchmarks to realistic large language model (LLM) workloads.

\paragraph{Small and Medium Matrices.}
The smallest configurations (e.g., $T=2048$, $D=128$ and $T=16384$, $D=256$) serve two purposes. First, they validate correctness and numerical stability under light workloads, where kernel launch overhead and instruction latency dominate. Second, they allow us to observe how different kernel designs behave before GPU parallelism is fully saturated. In these regimes, performance differences between naive, tiled, and coarsened kernels are intentionally modest, helping confirm that optimizations do not introduce regressions at small scale.

\paragraph{Large and Very Large Matrices.}
Configurations such as $T=65536$ and $T=131072$ with moderate head dimensions ($D=256$) are representative of intermediate transformer layers operating on long sequences. These tests are large enough to amortize kernel launch overhead and stress global memory bandwidth, making them ideal for comparing memory access patterns across kernel variants. This regime is particularly useful for validating whether shared memory tiling or coarsening provides any benefit over a well-coalesced naive implementation.

\paragraph{Realistic LLM Workloads.}
The largest test cases ($T=131072$ with $D \in \{1024, 2048, 4096, 8192\}$) are chosen to closely mirror KV-cache sizes encountered in modern large language models during long-context inference. These configurations approach billions of elements and represent the regime where quantization overhead must be negligible relative to attention computation to be practical. Evaluating performance here allows us to draw conclusions about whether GPU-based quantization is viable in production inference pipelines.

\paragraph{Motivation for Dimensional Scaling.}
We vary the head dimension $D$ independently of sequence length $T$ to study how performance and numerical error scale with model width. This is especially important for analyzing attention surrogate error, which depends directly on $D$. By sweeping $D$ across powers of two up to 8192, we ensure compatibility with vectorized kernels (e.g., \texttt{float4}) while also capturing realistic transformer design choices.

\paragraph{Benchmarking Philosophy.}
Finally, these test cases support two complementary evaluation perspectives. First, kernel-only timing isolates architectural effects such as memory coalescing, instruction-level parallelism, and vectorized loads. Second, end-to-end comparisons against a CPU baseline contextualize these gains in terms of practical speedups. Together, this test suite enables us to rigorously validate prior assumptions about tiling, coarsening, and vectorization under both controlled and realistic conditions.

\paragraph{Unit Testing and Validation}

To ensure correctness and robustness of both the CPU and GPU implementations, we developed a comprehensive unit testing framework covering functionality, numerical accuracy, and edge cases. The test suite consists of 25 individual tests and is executed independently of the benchmarking harness.

Basic structural tests verify correct matrix allocation, dimension handling, and memory initialization for both FP32 and INT8 matrix types. Randomized fill routines are validated to ensure values remain within specified bounds, preventing silent data corruption at later stages.

Numerical correctness is tested using identity checks for L2 error, maximum absolute error, and attention score error, all of which must evaluate to zero when comparing a matrix against itself. Additional deterministic tests validate scale computation, quantization rounding behavior, and dequantization accuracy on hand-constructed inputs where expected outputs are analytically known.

GPU kernels are validated by direct comparison against the CPU reference implementation. For each kernel variant (naive, tiled, coarsened, and vectorized), quantized outputs are compared element-wise to the CPU result, allowing a tolerance of $\pm1$ to account for minor rounding differences between CPU and GPU floating-point behavior. Cross-kernel consistency tests further verify that all GPU variants produce identical outputs for the same inputs.

Edge cases such as $1\times1$ matrices, minimal vectorizable dimensions, and structured input patterns (all zeros, all ones, alternating signs) are explicitly tested to ensure correctness under degenerate conditions. Finally, stress tests on larger matrices confirm stability and correctness under non-trivial workloads.

This validation framework provides strong confidence that all reported performance results correspond to correct and numerically consistent behavior, and that observed differences between kernel variants arise from architectural and memory-access characteristics rather than implementation errors.

\section{Conclusion}

This work implemented and evaluated GPU-accelerated INT8 quantization for KV cache compression in large language models. We developed a CPU baseline and four CUDA kernel variants---naive, tiled, coarsened, and vectorized---and evaluated them across realistic LLM workload sizes.

\textbf{Performance}: The vectorized kernel achieves up to 1,694$\times$ speedup over the CPU baseline, completing quantization of 1 billion elements in under 50ms.

\textbf{Memory reduction}: INT8 quantization provides 4$\times$ memory savings compared to FP32 storage, enabling longer context windows and larger batch sizes.

\textbf{Accuracy}: Maximum per-element error is bounded at 0.004, and attention score error remains below 0.1 even for 8K-dimensional heads---demonstrating minimal impact on model behavior.

\textbf{Practical viability}: For realistic LLM workloads, quantization overhead is just 6--58ms, making it negligible compared to actual inference time.

Our analysis reveals that KV cache quantization is fundamentally memory-bound, and vectorization provides the most effective optimization by reducing memory transaction overhead. These findings provide practical guidance for implementing efficient quantization in production LLM inference systems.

\subsection{Limitations}

This work focuses on per-channel INT8 quantization and does not explore several promising directions:

\begin{itemize}
    \item \textbf{Lower bit-widths}: INT4 or INT2 quantization could provide 8$\times$ or 16$\times$ compression at the cost of increased error.
    \item \textbf{Mixed-precision strategies}: Selective quantization of less critical layers or tokens could balance memory savings with accuracy.
    \item \textbf{Dynamic quantization}: Adaptive scale computation during inference could improve precision for out-of-distribution inputs.
    \item \textbf{End-to-end evaluation}: We measure reconstruction and attention score error but do not evaluate impact on downstream task performance (e.g., perplexity on language modeling benchmarks).
    \item \textbf{Multi-GPU scaling}: Our implementation targets single-GPU scenarios and does not address distributed inference workloads.
\end{itemize}
\subsection{Future Work}

Several promising directions extend this work toward more comprehensive optimization and real-world deployment evaluation. First, exploring additional GPU optimization strategies could further improve quantization performance. Warp-level primitives such as \texttt{\_\_shfl\_down\_sync} could enable more efficient reduction operations during scale computation, while persistent kernels that maintain thread blocks across multiple quantization operations could amortize launch overhead for streaming workloads. Additionally, investigating GPU-specific numeric formats such as FP8, which recent architectures natively support, could provide hardware-accelerated alternatives to INT8 with potentially better accuracy-performance tradeoffs.

Beyond kernel-level optimizations, the most critical next step is evaluating these quantization techniques within complete transformer blocks during actual inference. This requires integrating our kernels into existing inference engines such as vLLM, TensorRT-LLM, or HuggingFace Transformers, then measuring end-to-end latency, throughput, and memory consumption on real workloads. Such integration would reveal how quantization overhead interacts with attention computation, feedforward layers, and memory bandwidth contention in production scenarios. Furthermore, testing on diverse models ranging from smaller variants like GPT-2 to larger models approaching 70B parameters would establish whether our performance characteristics hold across different architectural scales and whether quantization enables previously infeasible batch sizes or context lengths.

Finally, comprehensive accuracy evaluation on standard benchmarks remains essential. While our attention score error analysis suggests minimal impact, measuring perplexity degradation on language modeling tasks, accuracy on question-answering benchmarks, and quality on generation tasks would quantify the practical tradeoff between memory savings and model capability. This end-to-end evaluation across GPU optimizations, inference system integration, and task-specific metrics would provide the complete picture necessary for deploying quantized KV caches in production LLM serving systems.

\section*{Acknowledgments}

We would like to thank Mohnish Sonsare for his contributions to this work, and Professor James Mooney for his guidance and support throughout this project.

\bibliographystyle{plain}

\end{document}